\documentclass[letterpaper, 10 pt, conference]{ieeeconf}  

\IEEEoverridecommandlockouts                              

\usepackage{graphics} 
\usepackage{epsfig} 
\usepackage{mathptmx} 
\usepackage{times} 
\usepackage{amsmath} 
\usepackage{amssymb}  
\usepackage{amsfonts}
\usepackage{booktabs}
\usepackage{multicol}
\usepackage{multirow}
\usepackage{color}

\title{\LARGE \bf
Fast and Incremental Loop Closure Detection \\Using Proximity Graphs
}

\author{Shan An$^{1}$, Guangfu Che$^{1}$, Fangru Zhou$^{1}$, Xianglong Liu$^{2,*}$, Xin Ma$^{3}$ and Yu Chen$^{1}$
\thanks{$^{}$This work was supported in part by the Chinese National Key Research and Development Plan (2018YFB1305803), Chinese National Natural Science Foundation (61673245), Chinese National Programs for High Technology Research and Development (2015AA042307).}%
\thanks{$^{1}$Shan An, Guangfu Che, Fangru Zhou and Yu Chen are with AR/VR department, JD.com, Beijing, China
        {\tt\small \{anshan, cheguangfu1, zhoufangru, chenyu6\}@jd.com}}
\thanks{$^{2}$Xianglong Liu is with School of Computer Science and Engineering, Beihang University, Beijing, China
       {\tt\small  xlliu@buaa.edu.cn}}%
\thanks{$^{3}$Xin Ma is with School of Control Science and Engineering, Shandong University, Jinan, China
       {\tt\small  maxin@sdu.edu.cn}}
\thanks{$^{*}$Corresponding Author}
}

\begin{document}

\maketitle
\thispagestyle{empty}
\pagestyle{empty}

\begin{abstract}

Visual loop closure detection, which can be considered as an image retrieval task, is an important problem in SLAM (Simultaneous Localization and Mapping) systems. The frequently used bag-of-words (BoW) models can achieve high precision and moderate recall. However, the requirement for lower time costs and fewer memory costs for mobile robot applications is not well satisfied. In this paper, we propose a novel loop closure detection framework titled `FILD' (Fast and Incremental Loop closure Detection), which focuses on an on-line and incremental graph vocabulary construction for fast loop closure detection. The global and local features of frames are extracted using the Convolutional Neural Networks (CNN) and SURF on the GPU, which guarantee extremely fast extraction speeds. The graph vocabulary construction is based on one type of proximity graph, named Hierarchical Navigable Small World (HNSW) graphs, which is modified to adapt to this specific application. In addition, this process is coupled with a novel strategy for real-time geometrical verification, which only keeps binary hash codes and significantly saves on memory usage. Extensive experiments on several publicly available datasets show that the proposed approach can achieve fairly good recall at 100\% precision compared to other state-of-the-art methods. The source code can be downloaded at \textit{https://github.com/AnshanTJU/FILD} for further studies.

\end{abstract}

\section{INTRODUCTION}

A mobile robot should have the ability of exploring unknown places and constructing the reliable map of environment while simultaneously using the map for the autonomous localization. The task is defined as the Simultaneous Localization And Mapping (SLAM) \cite{durrant2006simultaneous, bailey2006simultaneous}, which is one of the most central topics in robotics research. In SLAM, one major problem is Loop Closure Detection (LCD), that is, the robot must determine whether it has returned to a previously mapped area. With the increase in computing power, the mobile robots not only use range and bearing sensors such as laser scanners \cite{gutmann1999incremental}, radars and sonars \cite{tardos2002robust}, but also use single cameras \cite{cummins2008fab} or stereo-camera rigs \cite{engel2015large}. Exploiting the appearance information of a scene to detect previous visited places is called Visual Loop Closure Detection \cite{angeli2008fast, galvez2012bags, tsintotas2018assigning}.

\begin{figure}[t]
\centering
 \includegraphics[width=0.4\textwidth]{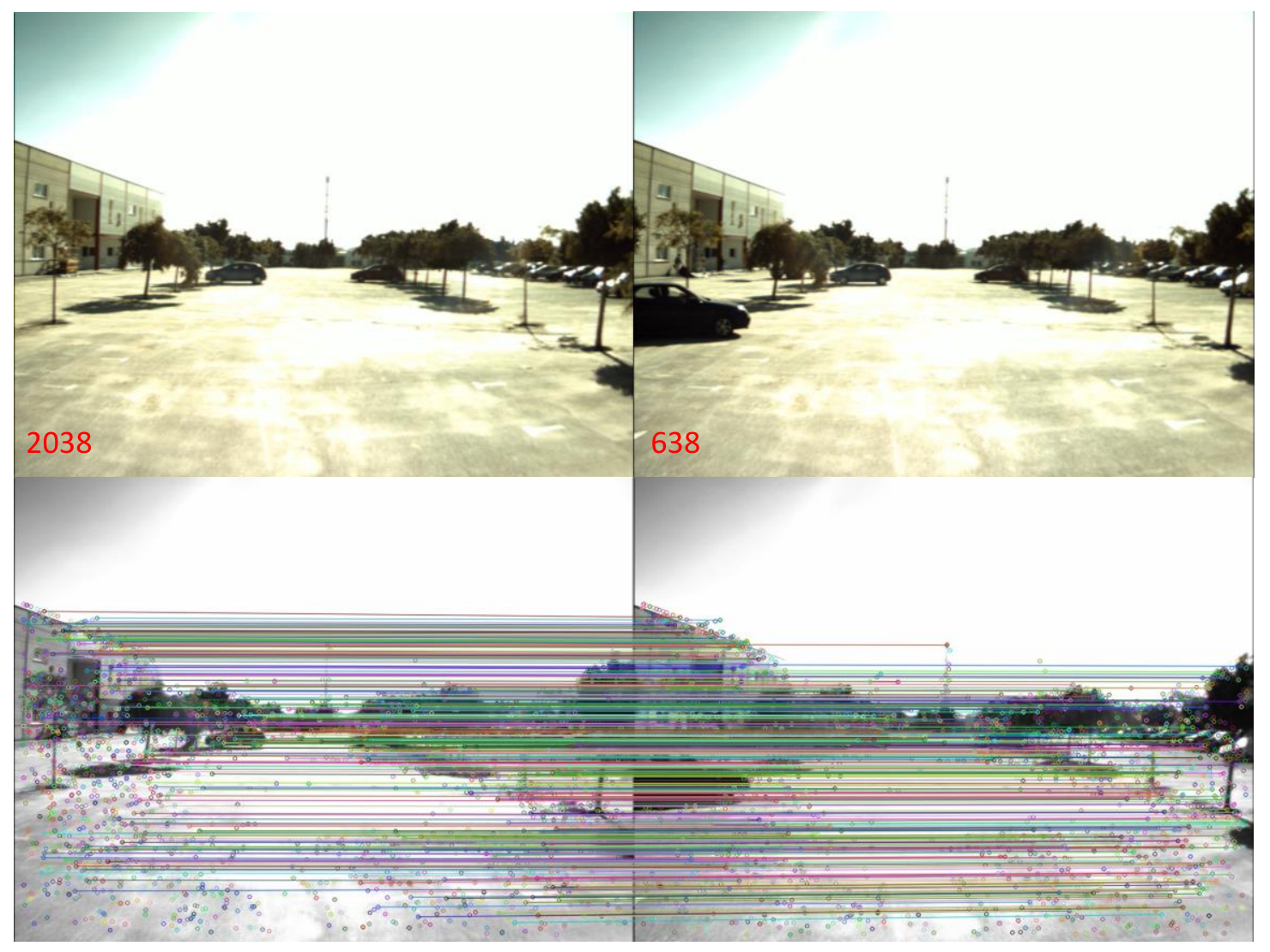}
\caption{The representation of image matching using CasHash \cite{cheng2014fast} and the proposed binary ratio test on Malaga 2009 Parking 6L \cite{blanco2009collection} dataset. (Top Left) The query image captured by the robot. (Top Right) The loop closure image which is returned by our system. (Bottom) The matches of two images are shown, which passed the binary ratio test and the RANSAC algorithm.}
\label{fig:figmatches}
\end{figure}

The visual loop closure detection problem can be converted into an on-line image retrieval task to determine if the current image has been taken from a known location. Conventional methods quantize the descriptor space of local features into Visual Words (VW), whether floating-point features, such as SIFT \cite{lowe2004distinctive}, SURF \cite{bay2006surf} or binary features, such as BRIEF \cite{calonder2010brief}, ORB \cite{rublee2011orb}. The so called BoW \cite{sivic2003video} employs the widely used ¡°term frequency-inverse document frequency¡± (tf-idf) technique to create a VW histogram. Pre-visited areas can be identified based on voting techniques \cite{gehrig2016visual} for place recognition.

The Convolutional Neural Networks (CNN) are designed to benefit and learn from massive amounts of data, which has demonstrated high performance in image classification \cite{krizhevsky2012imagenet} and scene recognition \cite{zhou2014learning}. Recently, with the outstanding discrimination power of CNN features, the landmarks in images are detected and matched for visual place recognition \cite{sunderhauf2015place}, which achieves better recognition accuracy than local features because of their invariance to illumination and their high-level semantics.

In this paper, we present a novel algorithm to detect loop closure, which is real-time and scalable, with the database built on-line and incrementally. Our approach is based on both the CNN features and SURF features, and using one type of proximity graph, named Hierarchical Navigable Small World (HNSW) graphs \cite{malkov2018efficient}. Several important novelties have been proposed, which make our algorithm much faster than current approaches. The images captured along the trajectory of the mobile robot is firstly described using the features of the pre-trained CNN. These features are used to construct the HNSW graphs by adding them into the graphs, and later they will be retrieved to get the top nearest neighbors according to image similarity. Finally, the geometrical consistency is confirmed using SURF features matched by CasHash \cite{cheng2014fast} and RANSAC. The main contributions of this paper are summarized as follows:

\begin{itemize}
\item A framework which uses CNN features and Hierarchical Navigable Small World graphs \cite{malkov2018efficient} to enable the incremental construction of the searching index and offer extremely fast on-line retrieval performance.
\item A novel strategy for real-time geometrical verification, with the important feature of using Hamming distances instead of Euclidian distances to perform the ratio test. The system only keeps binary hash codes instead of float-point descriptors, which will significantly save memory usage.
\item The source code of our implementation will be released to academia to facilitate future studies.
\end{itemize}

The rest of the paper is organized as follows. In Section II, we summarize relevant prior research in loop closure detection. In Section III, the proposed algorithm is described in detail. Our experimental design and comparative results are presented in Section IV. Conclusions and future work are discussed in Section V.

\begin{figure*}[t]
\centering
 \includegraphics[width=1\textwidth]{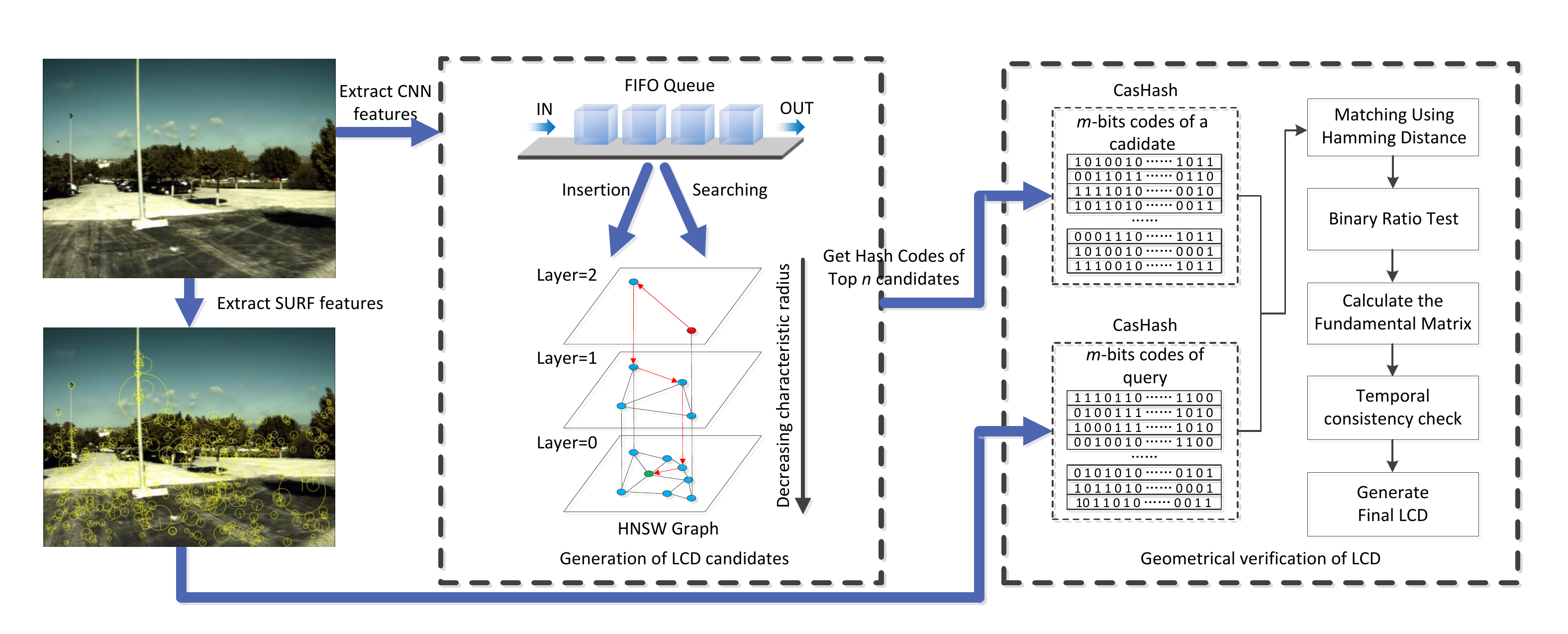}
\caption{An overview of the proposed loop closure detection method. As the incoming image stream enters the pipeline, the CNN features \cite{sandler2018mobilenetv2} and the SURF features \cite{bay2006surf} of the image are extracted. The CNN features enters the FIFO queue until the number of frames are more than $\psi \times \phi$, then the insertion into the HNSW graph \cite{malkov2018efficient} is performed. The searching of the HNSW graph will return top $n$ nearest neighbors and get the corresponding hash codes. Then the SURF features of the incoming image will convert to hash codes and using Hamming distance to perform matching. A binary ratio test is implemented to eliminate false matches, in conjunction with RANSAC to compute the fundamental matrix and generate final LCD.}
\label{fig:fighnsw}
\end{figure*}

\section{RELATED WORK}
\label{RW}
The methods for visual loop closure detection can be roughly divided into two classes: off-line and on-line. The off-line appearance-based FAB-MAP system \cite{cummins2008fab} and FAB-MAP 2.0 system \cite{cummins2011appearance} use a Chow Liu tree to learn a generative model of place appearance. A hierarchical BoW model with direct and inverse indexes built with binary features has been used to detect revisited places \cite{galvez2012bags}, with a geometrical verification step to avoid false positives. The sequences of images instead of single instances are represented by visual word histograms in \cite{bampis2016encoding}, and sequence-to-sequence matches are performed coherently advancing along time.

An on-line method \cite{angeli2008fast} using an incremental version of the BoW estimates the matching probability through a Bayesian filtering scheme. An incremental vocabulary building process proposed in \cite{nicosevici2012automatic} uses an agglomerative clustering algorithm. The stability of feature-cluster associations are increased using an incremental image-indexing process in conjunction with a tree-based feature-labeling method. The IBuILD system proposed in \cite{khan2015ibuild} uses an on-line and incremental formulation of binary vocabulary, with binary features between consecutive images being tracked to incorporate pose invariance and a likelihood function used to generate loop closures. In \cite{tsintotas2018assigning}, the incoming image stream is dynamically segmented to formulate ¡°places¡± and a voting scheme is used over the on-line generated visual words to locate the proper candidate place.

The methods above use local features such as SURF \cite{bay2006surf} and BRIEF \cite{calonder2010brief}. In early studies of place recognition, image representations are based on global descriptors, such as color or texture \cite{torralba2003context}. The global descriptors of images are evolved into CNN based features in recent years, which are used in the visual place recognition field \cite{sunderhauf2015performance} and loop closure detection \cite{hou2015convolutional}. However, using CNN features the robot could not get the topological information for the data association between the images, which is crucial for the SLAM algorithm. Therefore, in our system, we utilize the SURF feature for one to one image matching and geometrical verification, which serves as a complement of the CNN based global features.

The most frequently used image matching strategy in visual loop closure detection is the BoW model \cite{angeli2008fast,galvez2012bags,nicosevici2012automatic,khan2015ibuild}, or those that are enhanced using a tree structure, such as a hierarchical k-means tree \cite{galvez2012bags} or a k-d tree \cite{liu2012indexing}. Since the problem can be treated as an image retrieval problem, the traditional image retrieval methods such as Product Quantization (PQ) \cite{jegou2011product} and Hashing \cite{hou2018bocnf} could be used. A k-NN graph \cite{hajebi2014efficient} is constructed as the search index for the vocabulary, in which each visual word corresponds to a node in the graph. However, the search index is built over the visual words in an offline phase. HNSW graphs \cite{malkov2018efficient} have been shown to be powerful structures for approximate nearest neighbor search. This paper will investigate the ability of an on-line and incrementally graph building coupled with extremely fast computation speed of image similarities, which will be beneficial for loop closure detection.

\section{PROPOSED METHOD}

In this section a detailed description of the proposed LCD pipeline is presented. The algorithm leverages the GPU acceleration and HNSW graphs \cite{malkov2018efficient} to achieve real-time performance. The whole process can be summarized as two stages: the generation of LCD candidates and the verification of LCD.

In the first stage, a HNSW graph is built and retrieved using CNN features, which is extracted from the incoming frames. Using a First-in-First-out (FIFO) queue, the recently captured images can be filtered out in the retrieval process. We carefully choose a highly efficient CNN model to extract features, which has an extremely fast speed on GPU. The use of HNSW ensures the building and the retrieval process of the database cost a few milliseconds.

 In the second stage, SURF features are matched using CasHash \cite{cheng2014fast} matcher followed by ratio test \cite{lowe2004distinctive} and RANSAC to perform geometrical verification. We exploit the ratio test using Hamming distance instead of using the L2 distance of the original features, which will significantly save memory or disk space. The time-consuming process here is the extraction of SURF features. Therefore, we utilize GPU to accelerate it, which guarantee the high precision and rapid verification.

\subsection{Description of the Features}
The proposed loop closure recognition system utilizes a lightweight Deep Convolution Neural Network named MobileNetV2 \cite{sandler2018mobilenetv2}, which is based on an inverted residual structure with linear bottlenecks. MobileNetV2 allows very memory-efficient inferences which are suitable for mobile applications.

The CNN features are extracted using the final average pooling layer of MobileNetV2. The network architecture is simplified by merging the batch normalization layer with a preceding convolution \cite{awebsite}. The forwarding time will be decreased by adding this operation. The computational process can be written as:

\begin{equation}
\mathrm{\mathbf{\hat{f}}}_{i,j}=\mathrm{\mathbf{W}}_{BN}\cdot(\mathrm{\mathbf{W}}_{conv}\cdot\mathrm{\mathbf{f}}_{i,j}+\mathrm{\mathbf{b}}_{conv})+\mathrm{\mathbf{b}}_{BN}
\label{eq01}
\end{equation}

Here $\mathrm{\mathbf{W}}_{BN} \in{\mathbb{R}^{C\times C}}$ and $\mathrm{\mathbf{b}}_{BN}\in{\mathbb{R}^{C}}$ denote the weight matrix and bias of the normalized version $\hat{F}$ of a feature map $F$. The parameters of the convolution layer which precedes batch normalization are denoted as $\mathrm{\mathbf{W}}_{conv} \in{\mathbb{R}^{C\times (C_{prev}\cdot k^2)}}$ and $\mathrm{\mathbf{b}}_{conv} \in {\mathbb{R}^{C}}$, where $C_{prev}$ is the number of channels of the feature map $F_{prev}$ input to the convolutional layer and $k \times k$ is the filter size. A $k \times k$ neighborhood of $F_{prev}$ is unwrapped in to a $k^2\cdot{C_{prev}}$ vector ${\mathbf{f}}_{i,j}$. Then the batch normalization layer and the preceding convolution layer can be replaced by a single convolution layer with the following parameters:

\begin{equation}
\mathrm{\mathbf{W}} = \mathrm{\mathbf{W}}_{BN} \cdot \mathrm{\mathbf{W}}_{conv}
\label{eq02}
\end{equation}
\begin{equation}
\mathrm{\mathbf{b}} = \mathrm{\mathbf{W}}_{BN} \cdot \mathrm{\mathbf{b}}_{conv} + \mathrm{\mathbf{b}}_{BN}
\label{eq03}
\end{equation}

The local invariant feature used in our system is Speeded Up Robust Features (SURF) \cite{bay2006surf}, which is based on the Hessian matrix to find points of interest. Circular regions around the interest points are constructed in order to assign a unique orientation and thus gain invariance to image rotations. In order to achieve higher accuracy, the proposed algorithm utilizes the full SURF space, which is 128 dimensions.

\subsection{Generation of LCD candidates}

When the robot travels on the road, the camera mounted on it will capture images and extract CNN features using MobilenetV2 \cite{sandler2018mobilenetv2}. Then these features are used to build the HNSW graphs and perform the retrieval to generate loop closure candidates.

The similarity between the features is calculated using the normalized scalar product (cosine of the angle between vectors) \cite{sivic2007efficient}:
\begin{equation}
s_{pq}=\frac{X_p^T\cdot{X_q}}{{\left \| X_p \right \|}_2 \cdot {\left \| X_q \right \|}_2}
\label{eq04}
\end{equation}

Where $s_{pq}$ is the similarity score between images $I_p$ and $I_q$, and $X_p$ and $X_q$ are the CNN feature vectors corresponding to the images. ${\left \| X \right \|}_2=\sqrt{{X^T}X}$ is the $L_2$ norm of vector $X$.

 Our system employs a proximity graph approach, called HNSW graphs \cite{malkov2018efficient}, which outperforms the state-of-the-art approximate nearest neighbor search methods, such as tree based BoW \cite{muja2014scalable} models, PQ \cite{jegou2011product} and LSH \cite{andoni2015optimal}. In the following sub-sections we describe HNSW¡¯s properties and explain how to use HNSW to construct graph vocabulary and perform approximate nearest neighbor search with the strategy to filter out the recently captured images.

\subsubsection{Properties of HNSW}

The HNSW graph is a fully graph based incremental K-Nearest Neighbor Search (K-NNS) structure, as shown in Fig.~\ref{fig:fighnsw}. It is based on Navigable Small World (NSW) model \cite{kleinberg2000navigation}, which has logarithmic or polylogarithmic scaling of greedy graph routing. Such models are important for understanding the underlying mechanisms of real-life networks formation.

The graph $\boldsymbol{G=(V,E)}$ formally consists of a set of nodes (i.e. feature vectors) $\boldsymbol{V}$ and a set of links $\boldsymbol{E}$ between them. A link $e_{ab}$ connects node $a$ with node $b$, which is directed in HNSW. The neighborhood of a node $a$ is defined as the set of its immediately connected nodes.
HNSW uses strategies for explicit selection of the graph¡¯s enter-point node, separate links by different scales and selecting neighbors using an advanced heuristic. The links are separated according to their length scale into different layers and then search in a hierarchical multilayer graph, which allows a logarithmic scalability.

\subsubsection{Construction of Graph Vocabulary}
In a BoW model, the visual vocabulary is usually constructed using k-means clustering. A search index is built over the visual words, which are generated using feature descriptors extracted from a training dataset. The building of the vocabulary is off-line, which means that it is not flexible and can not adapt to every working environment.

HNSW has the property of incremental graph building \cite{malkov2018efficient}. The image features can be consecutively inserted into the graph structure. An integer maximum layer $l$  is randomly selected with an  exponentially decaying probability distribution for every inserted element. The insertion process starts  from the top layer to the next layer, by greedily traversing the graph in order to find the $ef$ closest neighbors to the inserted element $q$ in the layer. The founded closest neighbors from the previous layer will be used as an enter  point to the next layer. A greedy search algorithm is used to find closest neighbors in each layer. The process repeats until the connections of the inserted elements are established on the zero layer. In each layer higher than zero, the maximum number of connections that an element can have per layer is defined by the parameter $M$, which is the only meaningful construction parameter.

During the movement of the mobile robot, the CNN features of the images are inserted into the graph vocabulary. The whole process is on-line and incremental,  thus eliminating the need for prebuilt data. Therefore, the use of HNSW ensures the robot's working in various environment.

\subsubsection{K-NN Search and Adaption for LCD}
The K-NN Search algorithm is roughly equivalent to the insertion algorithm for an item with layer $l=0$, with the difference that the closest neighbors found at the ground layer are returned as the search result. The search quality is controlled by the parameter $ef$.

The images are captured sequentially and the adjacent images may have high similarities, which will result in false-positive LCDs. Therefore, we design a First-in-First-out (FIFO) queue to store image features. The image feature $X_q$ of the current Image $I_q$  is first inserted into the queue $Q$ and until the robot runs out of the search area the feature will be inserted into the HNSW graph. The search area that rejects recently acquired input frames is defined based on a temporal constant $\psi$ and the frame rate of the camera $\phi$. If the frames feed into the queue $Q$ more than $\psi \times \phi$, the insertion into the HNSW graph is performed, otherwise, it will only insert into the queue $Q$. Consequently, when we use the current feature as the query feature, it will only search in $N - \psi \times \phi$ database, where $N$ is the number of entire images up to now. The features in the search area will never appear in the results.

\subsection{Geometrical verification of LCD}
\label{GV}

\begin{table*}[]
\caption{The Descriptions of the Datasets}
\label{table_dataset}
\begin{center}
\begin{tabular}{c|c|c|c|c|c}
\toprule
Dataset & Description & Camera Position & Image Resolution & \# Images & Frames Per Second \\
\midrule
KITTI 00 \cite{fritsch2013new} & Outdoor, dynamic & Frontal & $1241\times376$ & 4541 & 10 \\
KITTI 05 \cite{fritsch2013new} & Outdoor, dynamic & Frontal & $1241\times376$ & 2761 & 10 \\
Malaga 2009 Parking 6L \cite{blanco2009collection} & Outdoor, slightly dynamic & Frontal & $1024\times768$ & 3474 & 7 \\
New College \cite{smith2009new} & Outdoor, dynamic & Frontal & $512\times384$ & 52480 & 20 \\
\bottomrule
\end{tabular}
\end{center}
\end{table*}

Our system incorporates a geometrical verification step for discarding outliers by verifying that the two images of the loop closure satisfy the geometrical constraint. As said in Section~\ref{RW}, we utilize the local SURF feature for image matching between a query $q$ and the top $n$ nearest neighbors. For verification, the fundamental matrix is computed using RANSAC, and then, the data association between the images can be derived with no extra cost, which can be used for any SLAM algorithm.

Here, we use the CasHash \cite{cheng2014fast} algorithm for pairwise image matching. The initial purpose of CasHash is rapid image matching for 3D reconstruction. The features of images are mapped into binary codes from coarse to fine. It uses $L$ hashing tables which have $m$ bits, and then each feature $p$ is assigned to a bucket $g_l(p)$. The $L$ functions $g_l(q)$ are represented in Eqn.\ref{eq05}, where $h_{s,l}(1\leq s \leq m, 1 \leq l \leq L$  are generated independently and uniformly at random from a locality sensitive family $\mathcal{H}$:
\begin{equation}
g_l(q)=(h_{1,l}(q), h_{2,l}(q),\cdots, h_{m,l}(q)), l=1,2,\cdots,L.
\label{eq05}
\end{equation}

The original SURF feature has 128D float-point descriptors, while using the CasHash the features can be changed to binary codes with $m$ bits. In the traditional use of CasHash, a ratio test is performed using the full feature space. However, in the application for a mobile robot, the memory of the mounted computer of the robot is limited. The cost of saving all SURF features of all frames is not practical. We propose that using the binary codes instead of the full features for the ratio test. The binary ratio test threshold $\varepsilon$ is defined as:
\begin{equation}
d_h(C_a, C_b^1) / d_h(C_a, C_b^2) \leq {\varepsilon}^2
\label{eq06}
\end{equation}

Here $d_h(\cdot)$ indicates the Hamming distance computation. $C_a$ are the binary codes of the descriptor $f_a$ in an image $I_a$, while $C_b^1$ and $C_b^2$ are the binary codes of two closest descriptors $f_b^1$ and $f_b^2$ in an image $I_b$. The feature matches which have lower ratio than $\varepsilon^2$ will be treated as good matches and feed into the RANSAC process to estimated a fundamental matrix $T$ between the query and the loop closure candidate images. In Fig.~\ref{fig:figmatches}, a representation of the image matching is shown, which use the binary ratio test and RANSAC to remove outliers.

The loop closure candidate is ignored if it fails to compute $T$ or the number of inlier points between the two images is less than a parameter $\tau$. A temporal consistency check is incorporated to examine whether the aforementioned conditions are met for the $\beta$ consecutive camera measurements, which is the same as the method used in \cite{tsintotas2018assigning}.

After the CasHash, each feature is encoded as $m$ bit hashing codes. For example, if we use $m=128$, which equate to 128 bits in the hashing, the memory usage will decrease from 128 floats to 128 bits, which means it only cost $1/32$  memory. This feature is very important in mobile robot applications, which have less memory than the servers.

\section{EXPERIMENTAL EVALUATION}
The evaluation datasets contain four publicly available image sets: KITII 00 \cite{fritsch2013new}, KITTI 05 \cite{fritsch2013new},
Malaga 2009 Parking 6L \cite{blanco2009collection} and New College \cite{smith2009new}.
A more detailed description of the datasets can be seen in Table \ref{table_dataset}.
The ground truth of these datasets are provided by the authors in \cite{tsintotas2018assigning} and the authors in \cite{galvez2012bags}. The performance of our method is compared against the state-of-the-art methods such as, FAB-MAP 2.0 \cite{cummins2011appearance},IBuILD \cite{khan2015ibuild},Bampis et al. \cite{bampis2016encoding}, Gehrig et al. \cite{gehrig2016visual}, G$\rm \acute{a}$lvez-L$\rm \acute{o}$pez et al. \cite{galvez2012bags}, Tsintotas et al. \cite{tsintotas2018assigning}.

\subsection{Method Evaluation}

We train the CNN network using the Place365 \cite{zhou2018places} dataset, which has 10 million images and 365 classes of scene for scene recognition. The top 1 accuracy is 51.47\% and top 5 accuracy is 82.61\%. We use this model in the following experiments.

The parameters of our method include three parts: the parameters of SURF features, HNSW graph, and geometrical verification. We use the default parameters of SURF, because it is not the research emphasis of this paper. An implementation of the loop closure detection algorithm presented in this paper is distributed as an open source code. 

For HNSW graph construction and searching, there are two parameters that could affect the search quality: the number of nearest to $q$ elements to return, $ef$;  and the maximum number of connections for each element per layer, $M$.  The range of the parameter $ef$ should be within 200, because the increase in $ef$  will lead to little extra performance but in exchange, significantly longer construction time. The range of the parameter $M$ should be 5 to 48.  The experiments in \cite{malkov2018efficient} show that bigger $M$ is better for high recall and high dimensional data, which also defines the memory consumption of the algorithm.  The temporal constant $\psi$ using in the FIFO queue will be set to 40 seconds in the rest of the paper. For geometrical verification, the parameters are: the hashing bits $m$, the ratio for binary ratio test $\varepsilon$, and the returned number of nearest neighbors $n$. The inlier points threshold $\tau$ is set to 20 empirically.

\begin{figure}[ht]
\begin{center}
\begin{multicols}{2}
   \includegraphics[width=1.0\linewidth]{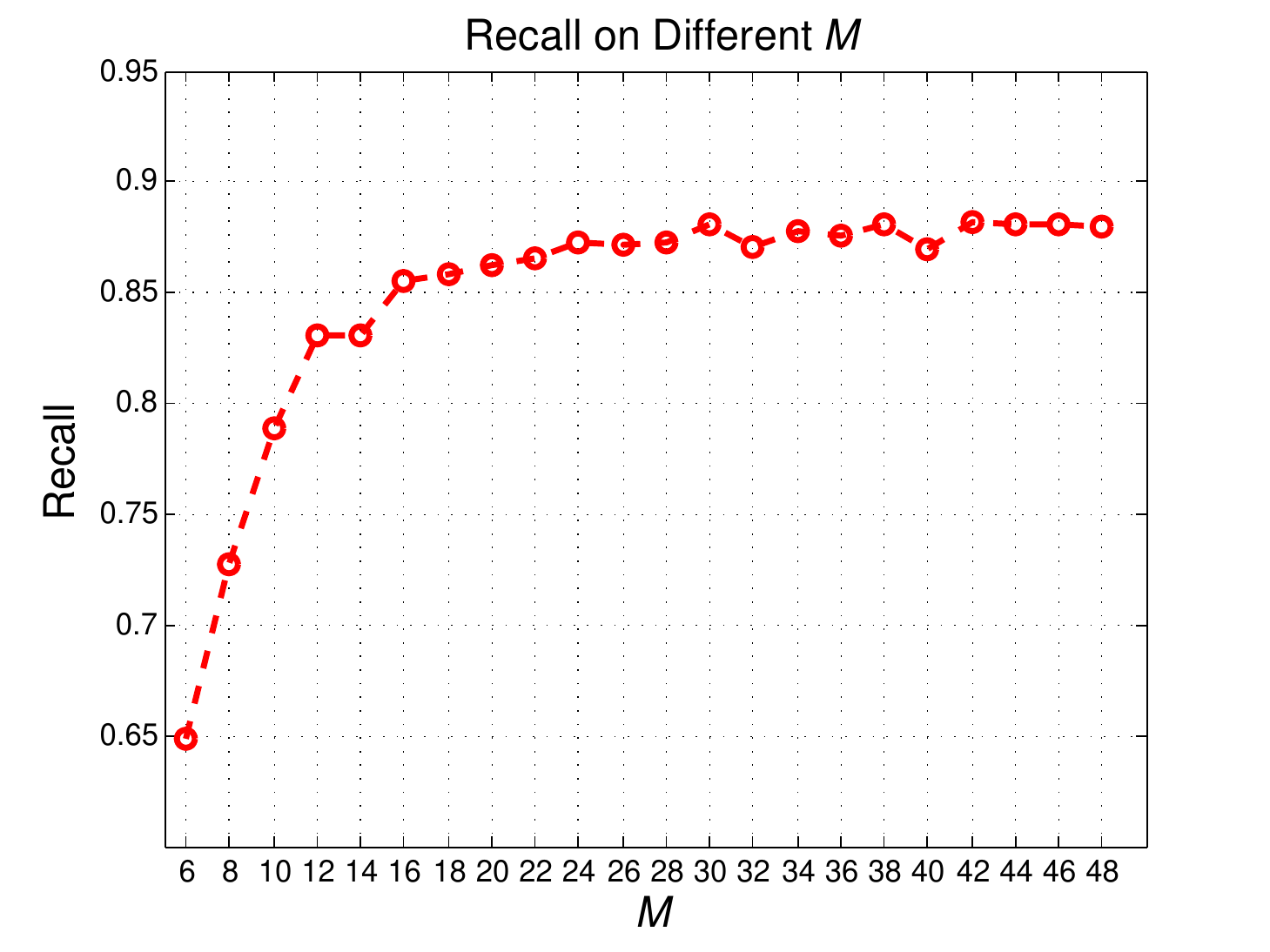} \par
   \includegraphics[width=1.0\linewidth]{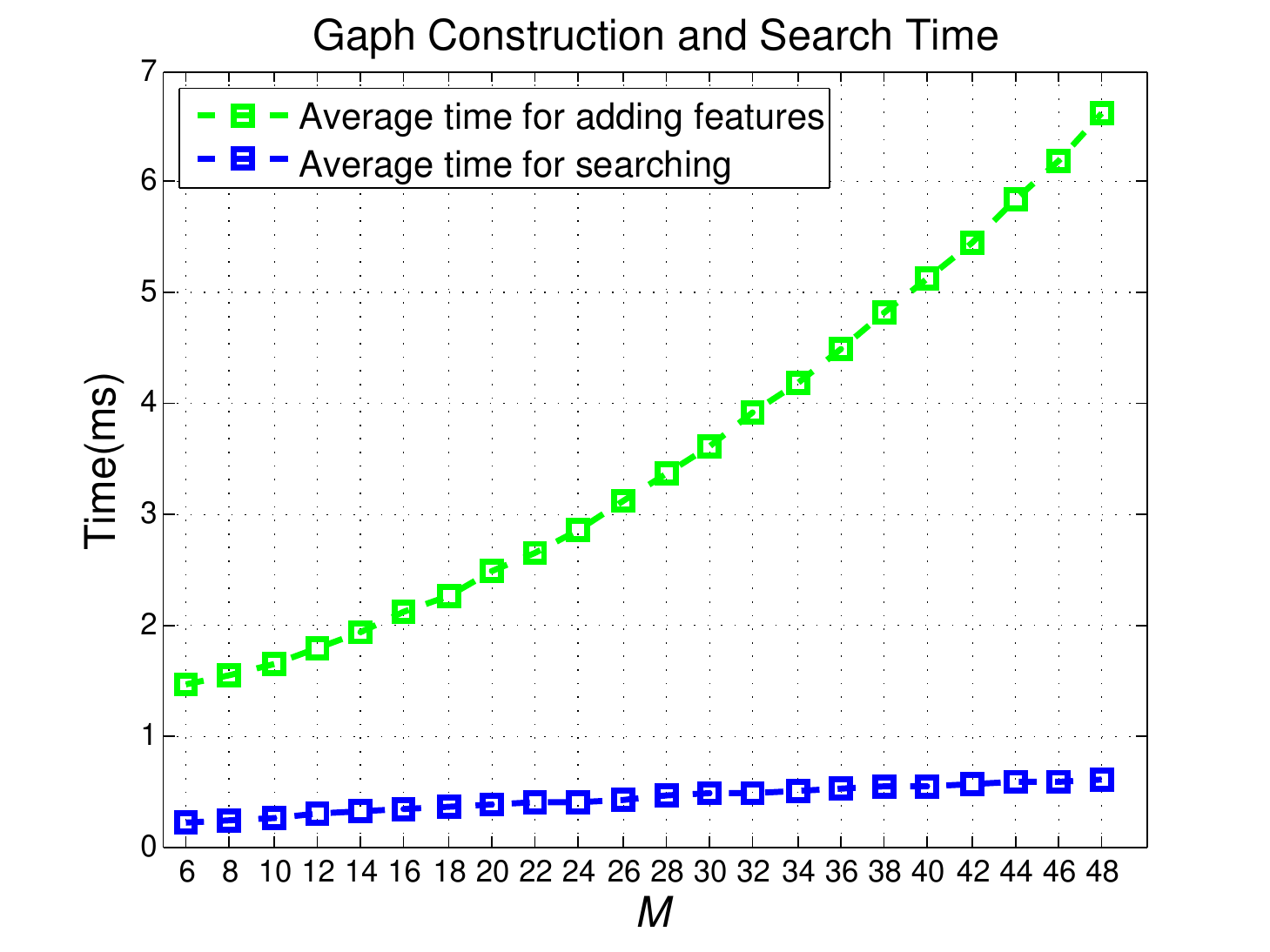} \par
 \end{multicols}
\end{center}
   \caption{(Left) The recall at 100\% precision of our algorithm on the New College \cite{smith2009new} dataset using a different $M$ from 6 to 48. (Right) The graph construction time and the searching time on the New College dataset using a different $M$.}
\label{fig:eval_M}
\end{figure}

\begin{figure}[ht]
\begin{center}
\begin{multicols}{2}
   \includegraphics[width=1.0\linewidth]{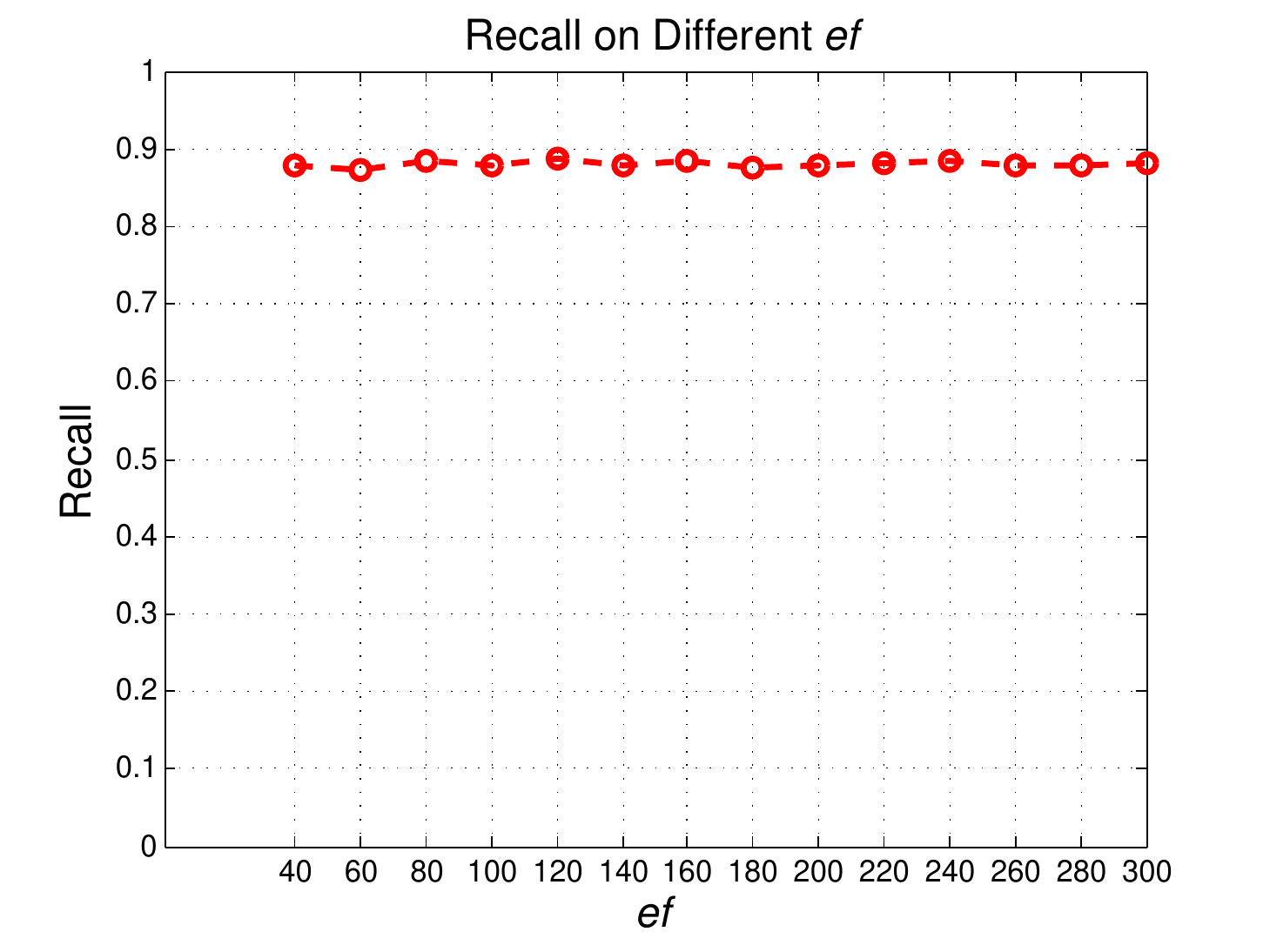} \par
   \includegraphics[width=1.0\linewidth]{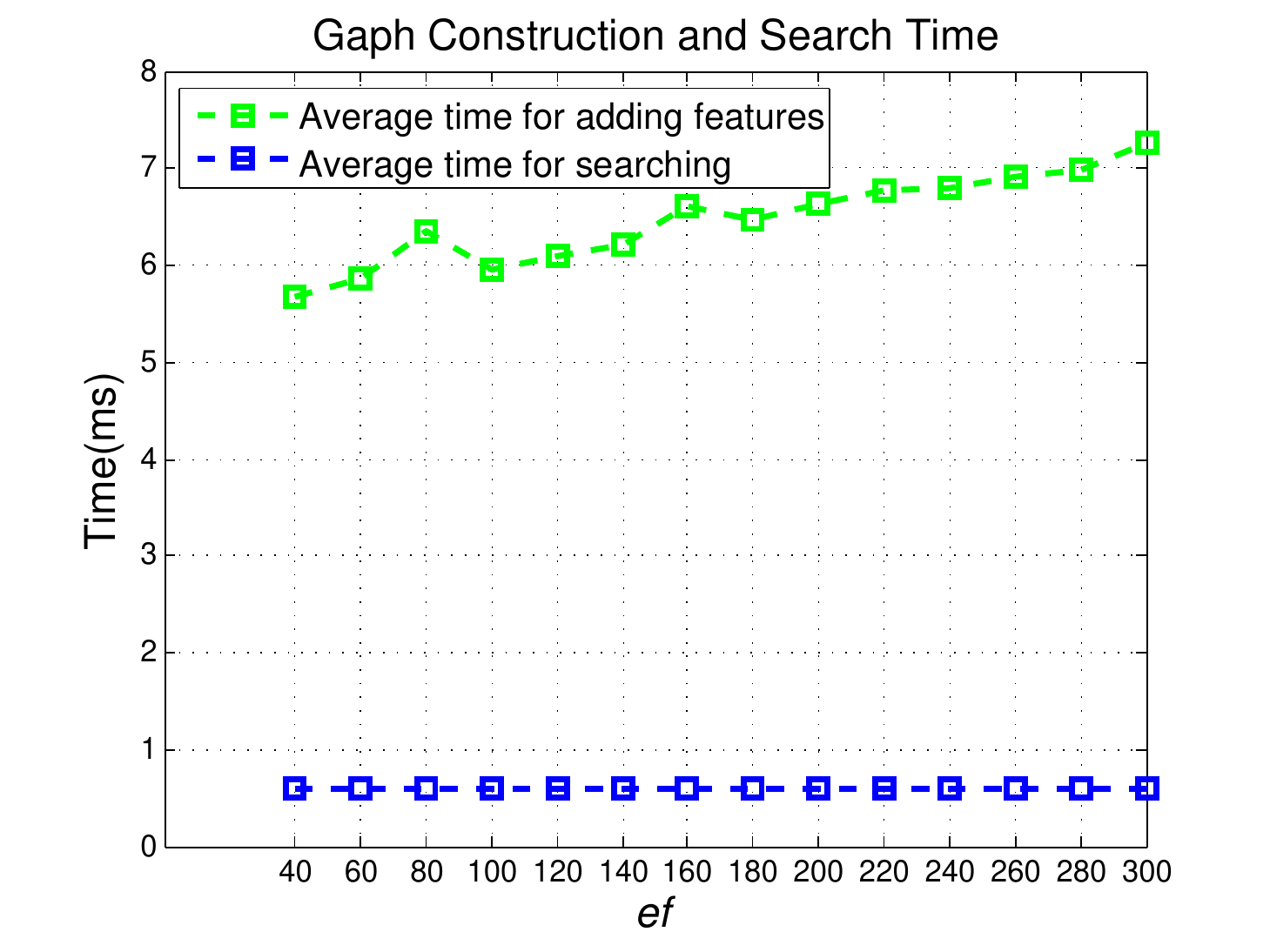} \par
 \end{multicols}
\end{center}
   \caption{(Left) The recall at 100\% precision of our algorithm on the New College \cite{smith2009new} dataset using a different $ef$ from 40 to 300. (Right) The graph construction time and the searching time on the New College dataset using a different $ef$.}
\label{fig:eval_EF}
\end{figure}
Firstly, we perform the experiments on the New College dataset\cite{smith2009new}  to choose $M$ and $ef$ for the HNSW graph retrieval. The other parameters are set as: $m=256$, $\varepsilon=0.7$. $ef$ is set to 200 when we change $M$. The returned number of nearest neighbors $n$ is set to 1. As 100\% precision can be reached with the temporal consistency check. The recalls are shown in the left part of Fig.~\ref{fig:eval_M}. We can see when $M$ increase, the recall will also increase. In the right part of Fig.~\ref{fig:eval_M}, the feature adding time and searching time will be increased when $M$ increases.

To evaluate different $ef$, the parameter $M$ is set to 16. It can be seen that in the left part of Fig.~\ref{fig:eval_EF}, the recall does not significantly change when the $ef$ increases. In the right part of Fig.~\ref{fig:eval_EF}, the feature adding time will be increased when $ef$ increases, while the searching time remains with no growth. According to the recall curve in Fig.~\ref{fig:eval_M} and Fig.~\ref{fig:eval_EF}, we chose $M=48$ and $ef=40$ in the following experiments.

\begin{figure}[ht]
\begin{center}
\begin{multicols}{2}
   \includegraphics[width=1.0\linewidth]{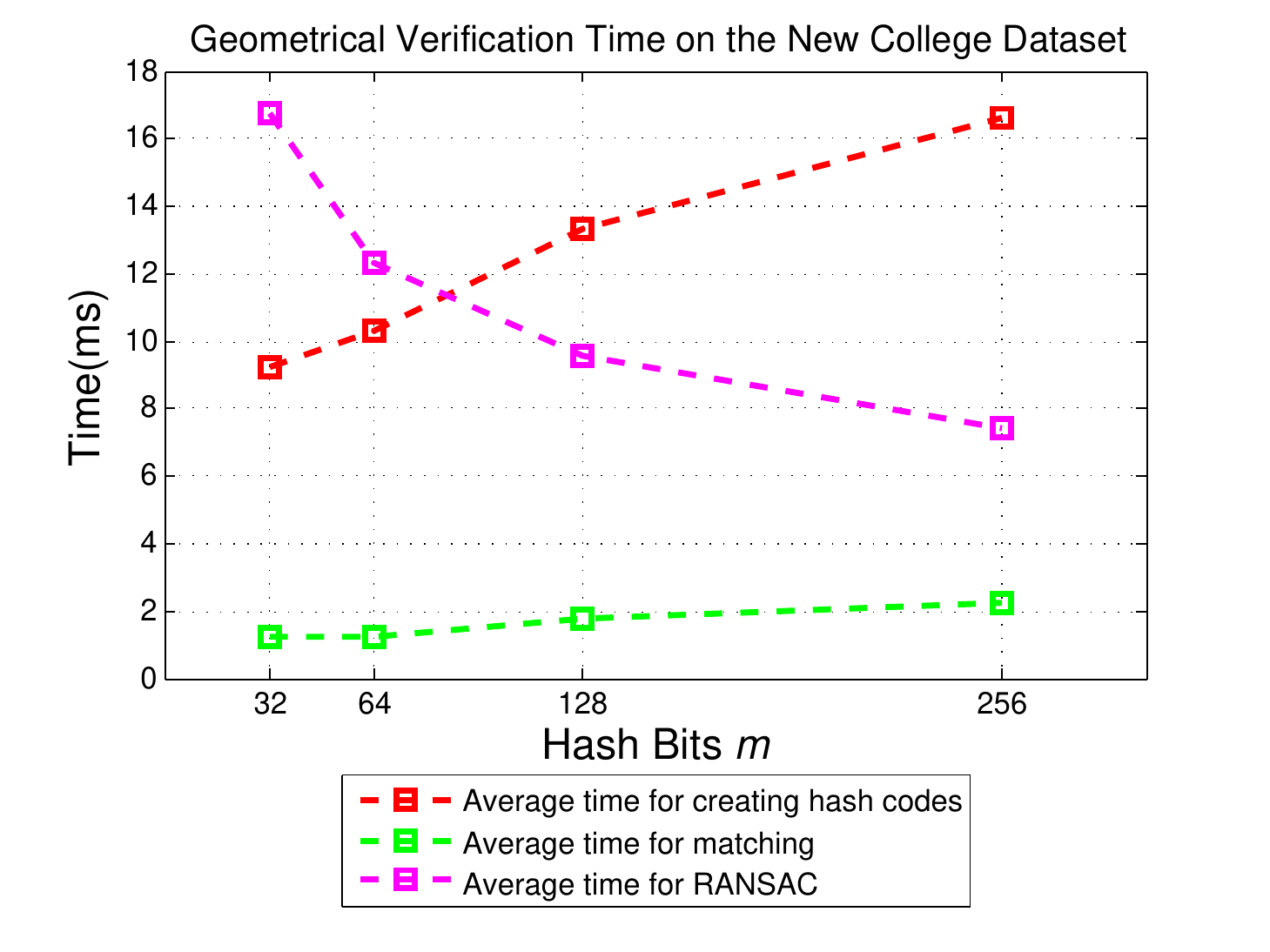} \par
   \includegraphics[width=1.0\linewidth]{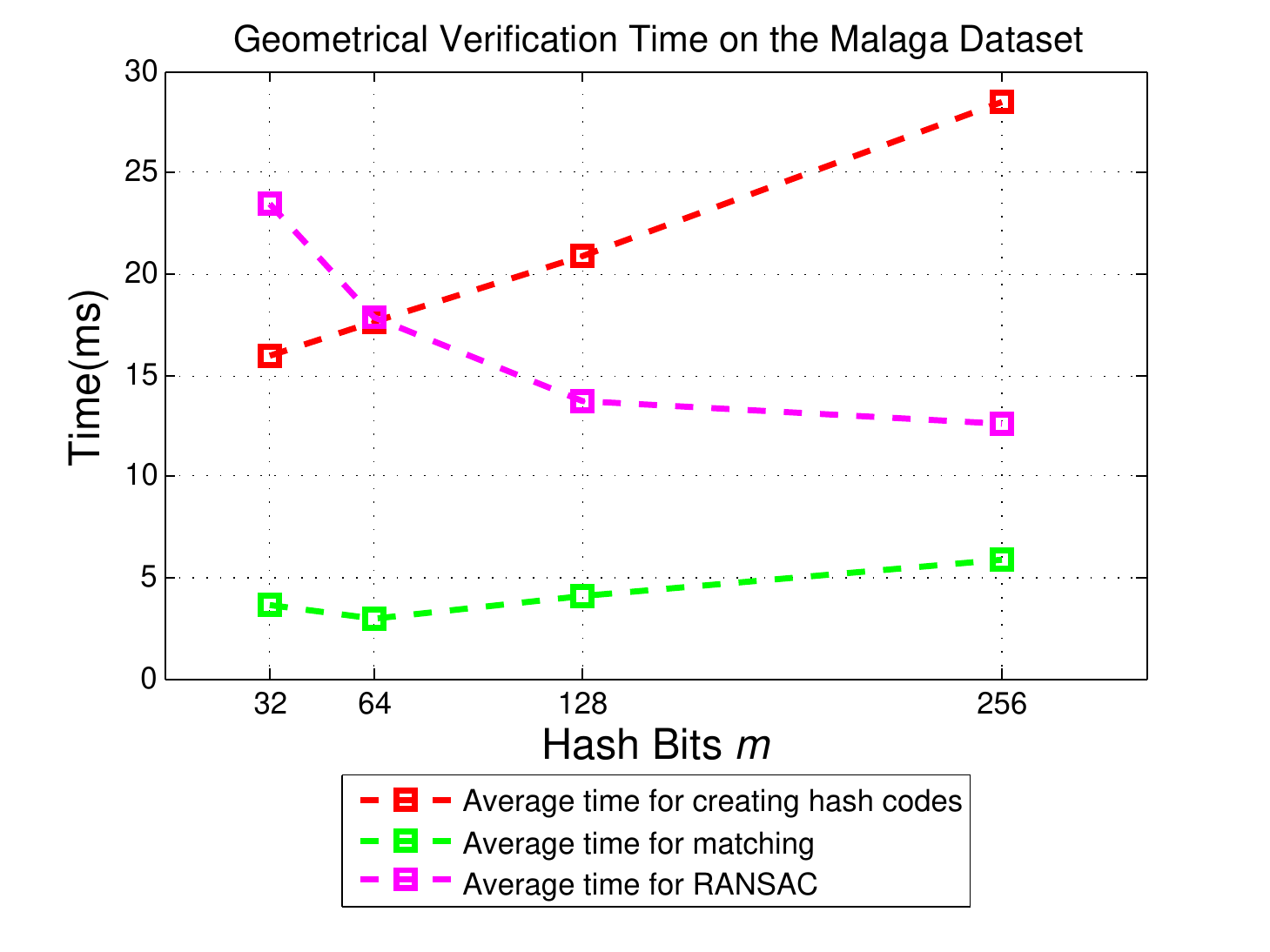} \par
 \end{multicols}
\end{center}
   \caption{The geometrical verification time on the New College dataset (Left) and the Malaga dataset (Right) using a different hashing bit $m$.}
\label{fig:eval_bits_time}
\end{figure}

Secondly, the hashing bits $m$ and the ratio $\varepsilon$ are evaluated. To evaluate the hash bits $m$, the ratio are set as $\varepsilon=0.7$. The returned number of nearest neighbors $n$ is set to 1. The temporal consistency check is incorporated in these experiments. The recalls of New College dataset and Malaga dataset are shown in Table~\ref{table_nc_nbits} and Table~\ref{table_ml_nbits}. It can be seen that using more hashing bits will increase the recall. In the Fig.~\ref{fig:eval_bits_time}, we can see that the hash codes creating time and the matching time will be increased when the hash bits $m$ increase, while the RANSAC time will be decreased. We chose $m=256$ in the remaining experiments, because the increase of the time is acceptable and the recall is better.

\begin{table}[]
\caption{The performance of New College Dataset with Different Hashing Bits $m$}
\label{table_nc_nbits}
\begin{center}
\begin{tabular}{c|c|c|c|c}
\toprule
Different Hashing Bits & 32 & 64 & 128 & 256 \\
\midrule
Recall (\%) & 87.83 & 88.30  & 89.34 & 90.67 \\
Precision (\%) & 100.0 & 100.0 & 100.0 & 100.0 \\
\bottomrule
\end{tabular}
\end{center}
\end{table}

\begin{table}[]
\caption{The performance of Malaga Dataset with Different Hashing Bits $m$}
\label{table_ml_nbits}
\begin{center}
\begin{tabular}{c|c|c|c|c}
\toprule
Different Hashing Bits & 32 & 64 & 128 & 256 \\
\midrule
Recall (\%) & 87.92 & 82.72  & 82.38 & 85.23 \\
Precision (\%) & 90.81 & 97.82 & 99.59 & 99.80 \\
\bottomrule
\end{tabular}
\end{center}
\end{table}

\begin{figure}[ht]
\begin{center}
\begin{multicols}{2}
   \includegraphics[width=1.0\linewidth]{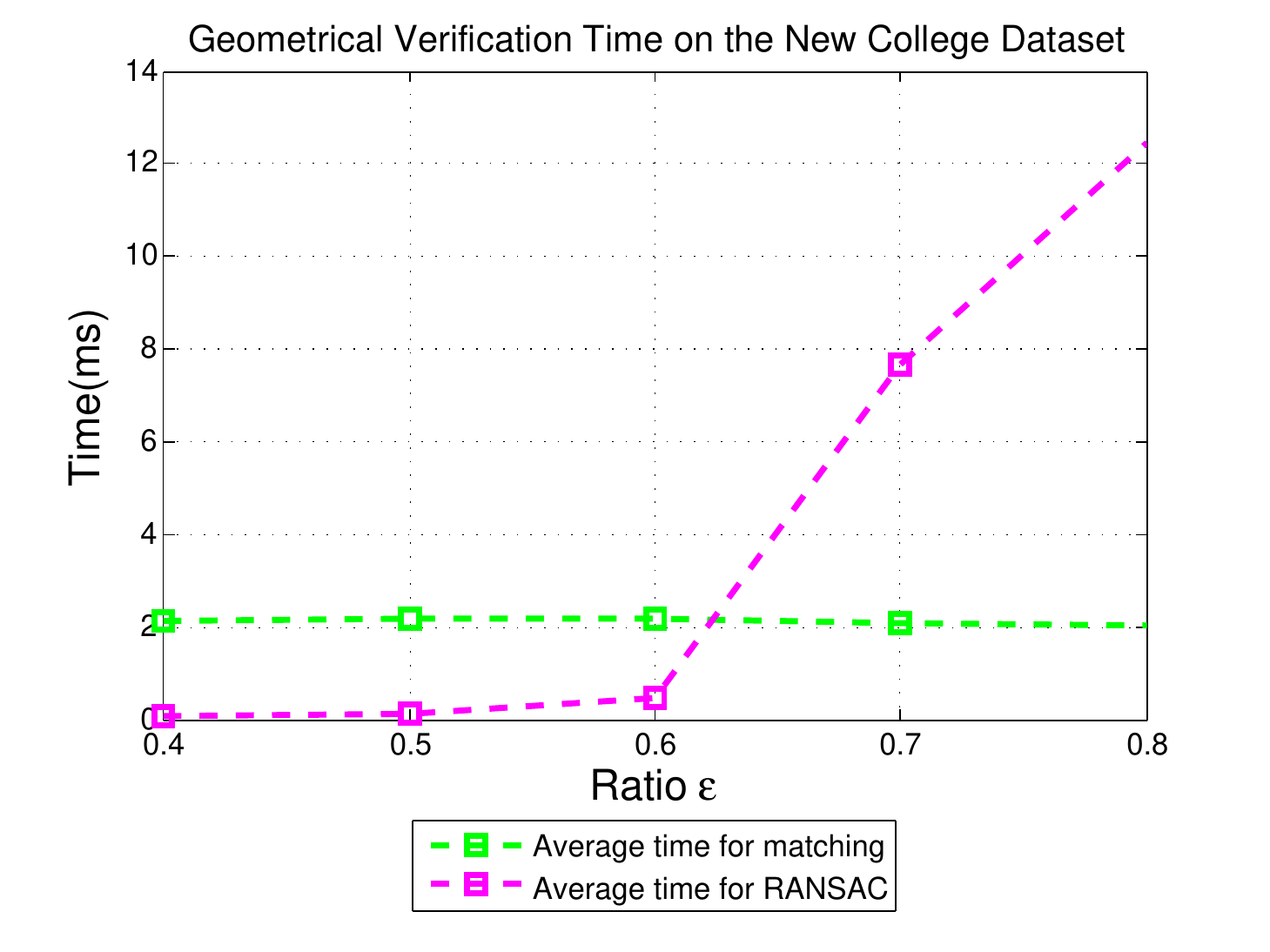} \par
   \includegraphics[width=1.0\linewidth]{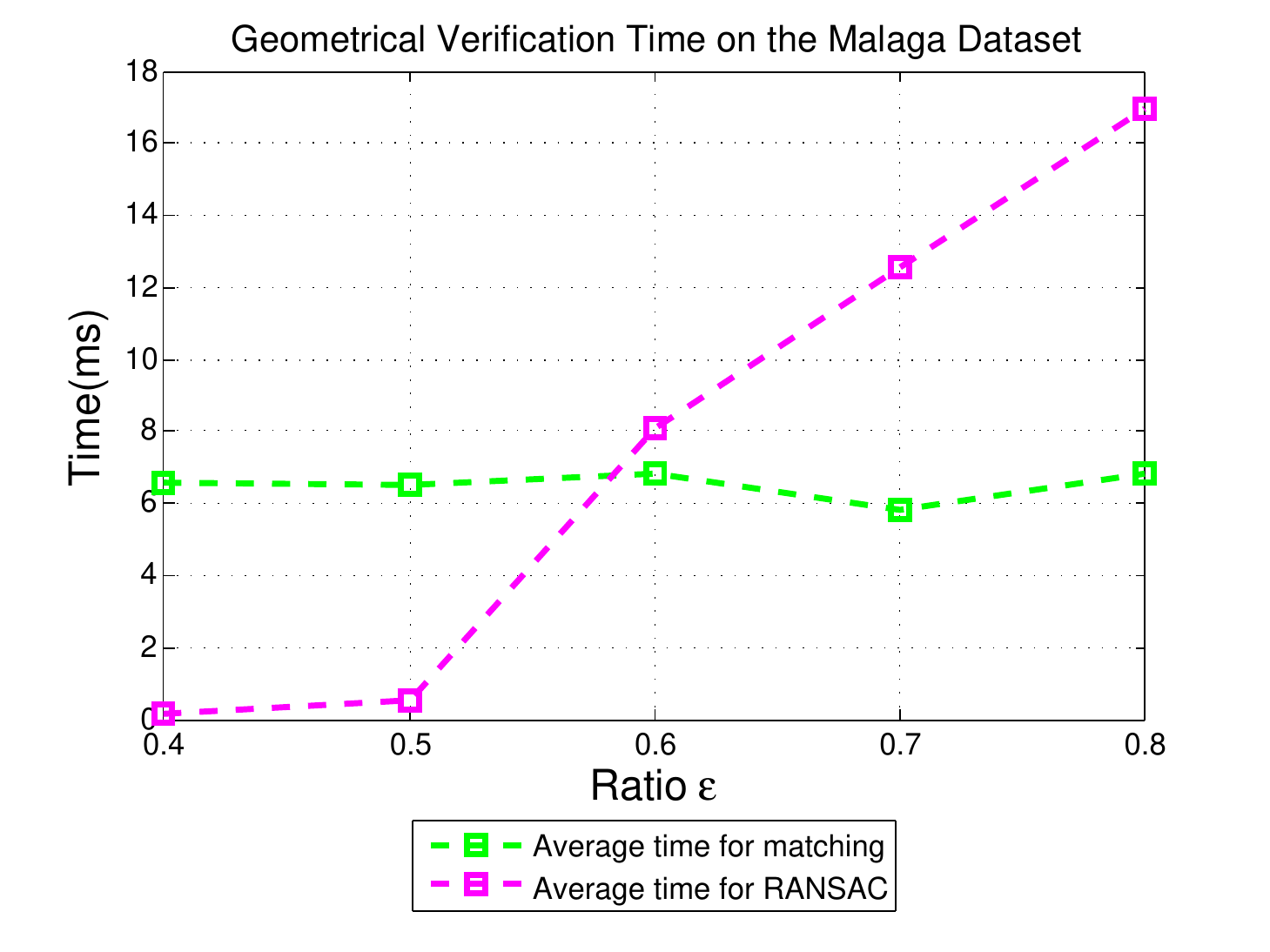} \par
 \end{multicols}
\end{center}
   \caption{The geometrical verification time on the New College dataset (Left) and the Malaga dataset (Right) using a different ratio of binary ratio test $\varepsilon$.}
\label{fig:eval_ratio_time}
\end{figure}

The ratio $\varepsilon$ of the binary ratio test is also very important for the precision and the recall of our system. We set $n=1$, and the temporal consistency check is used to evaluate the ratio. The recalls of New College dataset and Malaga dataset will increase as the ratio increases, as shown in Table~\ref{table_nc_ratio} and Table~\ref{table_ml_ratio}. The hash matching time will not increase during the change of the ratio, while the RANSAC time will be increased significantly, as shown in Fig.~\ref{fig:eval_ratio_time}. We chose $\varepsilon=0.7$ to ensure the precision to be 100\% and to achieve a higher recall.

\begin{table}[]
\caption{The performance of New College Dataset with Different Ratio $\varepsilon$}
\label{table_nc_ratio}
\begin{center}
\begin{tabular}{c|c|c|c|c|c}
\toprule
Ratio of Binary Ratio Test & 0.4 & 0.5 & 0.6 & 0.7 & 0.8 \\
\midrule
Recall (\%) & 30.84 & 57.57  & 78.42 & 88.73 & 92.35 \\
Precision (\%) & 100.0 & 100.0 & 100.0 & 100.0 & 100.0\\
\bottomrule
\end{tabular}
\end{center}
\end{table}

\begin{table}[]
\caption{The performance of Malaga Dataset with Different Ratio $\varepsilon$}
\label{table_ml_ratio}
\begin{center}
\begin{tabular}{c|c|c|c|c|c}
\toprule
Ratio of Binary Ratio Test & 0.4 & 0.5 & 0.6 & 0.7 & 0.8 \\
\midrule
Recall (\%) & 43.22 & 55.34  & 67.78 & 81.82 & 92.98 \\
Precision (\%) & 100.0 & 100.0 & 100.0 & 100.0 & 97.49 \\
\bottomrule
\end{tabular}
\end{center}
\end{table}

Finally, the returned number of nearest neighbors $n$ is evaluated. We can see in the Table ~\ref{table_nc_topn} and Table ~\ref{table_ml_topn}, the recall will be increased when the $n$ increased. For the Malaga dataset \cite{blanco2009collection}, the recall is 80.54\% at 100\% precision when it returned the nearest neighbor, while increasing $n$ will cause a decrease in precision. Because the 100\% precision is important for the loop closure detection, we selected $n=1$. Using more nearest neighbor in the geometrical verification stage will cost more time for hash code matching and RANSAC. Therefore, using only the nearest neighbor will bring a reduction in processing time. According to the above experiments, we determine the parameters of our algorithm, which are summarized in Table~\ref{table_parameter}.

\begin{table}[]
\caption{The performance of New College Dataset with Different Number Of Loop Closure Candidates $n$}
\label{table_nc_topn}
\begin{center}
\begin{tabular}{c|c|c|c|c|c|c}
\toprule
Nearest Neighbors & 1 & 2 & 4 & 6 & 8 & 10 \\
\midrule
Recall (\%) & 89.94 & 94.85  & 97.67 & 97.76 & 97.85 & 98.41 \\
Precision (\%) & 100.0 & 100.0 & 100.0 & 100.0 & 100.0 & 100.0\\
\bottomrule
\end{tabular}
\end{center}
\end{table}

\begin{table}[]
\caption{The performance of Malaga Dataset with Different Number Of Nearest Neighbors $n$}
\label{table_ml_topn}
\begin{center}
\begin{tabular}{c|c|c|c|c|c|c}
\toprule
Nearest Neighbors & 1 & 2 & 4 & 6 & 8 & 10 \\
\midrule
Recall (\%) & 80.54 & 89.19  & 97.95 & 96.69 & 97.33 & 96.49 \\
Precision (\%) & 100.0 & 99.82 & 99.36 & 99.23 & 99.24 & 99.25 \\
\bottomrule
\end{tabular}
\end{center}
\end{table}

\subsection{Comparative Results}

In the Table ~\ref{table_Comparativeresults}, the precision and recall of the proposed method against the aforementioned state-of-the-art methods are compared.  The best, second and third best results are marked in red, blue and green, respectively. Our method best in the New College dataset, 2 points higher than Tsintotas et al. \cite{tsintotas2018assigning}. In the Malaga dataset, our method achieved 80.54\% recall at 100\% precision, which is lower than Tsintotas' method \cite{tsintotas2018assigning} and was second best. For KITTI 00 and KITTI 05 dataset, our method was higher than Bampis's method \cite{bampis2016encoding}.

\subsection{Execution Time and Memory Usage}

We evaluated the feature extraction time on the GPU. The forwarding time of MobileNetV2 \cite{sandler2018mobilenetv2} was 13.33 ms, while the forwarding time of merging the batch normalization layer was 5.35 ms, which achieve an obvious speed acceleration.

To measure the execution time of whole system, we ran our system using the New College dataset \cite{smith2009new} using the parameters in Table~\ref{table_parameter}. The first experiment used the working frequency $f=1Hz$, which processed a total of 2624 images. The execution time of our system cost 48.73 ms per image on average and a peak of  83.70 ms. In order to test the scalability of the system, we set the frequency to $f=20Hz$ and obtained 52480 images. The execution time consumed per image in that case is shown in Table~\ref{table_timecost}. This was measured on a Intel(R) Xeon(R) CPU E5-2640 v4 @ 2.40GHz machine, with a NVIDIA P40 GPU card. The average running time per image was about 50 ms, which is very close to that using 2624 images and fast enough for loop closure detection. The average running time of Tsintotas's method \cite{tsintotas2018assigning} is about 300 ms, which is 6 times higher than our method.

\begin{table}[h]
\caption{Parameter List}
\label{table_parameter}
\begin{center}
\begin{tabular}{|c||c|}
\hline
Number of nearest to $q$ elements to return, $ef$ & 40 \\
\hline
Maximum number of connections for each element per layer, $M$ & 48 \\
\hline
Search area time constant, $\psi$ & 40 \\
\hline
Hashing bits, $m$ & 256 \\
\hline
Ratio of binary ratio test, $\varepsilon$ & 0.7\\
\hline
Geometrical verification inliers, $\tau$ & 20 \\
\hline
Images¡¯ temporal consistency, $\beta$ & 2 \\
\hline
Number of returned nearest neighbors, $n$ & 1 \\
\hline
\end{tabular}
\end{center}
\end{table}

As described in section ~\ref{GV}, we use CasHash \cite{cheng2014fast} for image matching, which quantize the SURF features into binary hashing codes. The proposed binary ratio test can avoid having to save the full float-point features. The memory usage of using full float-point features in our system is 28.11 GB, while using hashing codes only cost 18.99 GB, which saves 32\% of memory usage.

\begin{table}[]
\caption{Execution Time In New College Dataset With 52480 Images }
\label{table_timecost}
\begin{center}
\begin{tabular}{c|c}
\toprule
Stages  & Mean Time (ms/query)\\
\midrule
CNN Feature Extraction & 8.72 \\
\hline
SURF Feature Extraction & 8.97 \\
\hline
Hash Codes Creation & 16.94 \\
\hline
Adding CNN Feature & 5.21 \\
\hline
Graph Searching & 0.93 \\
\hline
Hash Codes Matching & 2.23 \\
\hline
RANSAC & 7.55 \\
\hline
Whole System & 50.28 \\
\bottomrule
\end{tabular}
\end{center}
\end{table}

\begin{table}[]
\caption{Comparative Results}
\label{table_Comparativeresults}
\begin{center}
\begin{tabular}{c|c|c|c}
\toprule
\tiny{Dataset} & \tiny{Approaches} & \tiny{Precision (\%)} & \tiny{Recall (\%)} \\
\midrule
\multirow{2}{*}{\tiny{KITTI 00 \cite{fritsch2013new}}} & \tiny{Gehrig et al. \cite{gehrig2016visual}} & \tiny{100} & \tiny{\textbf{\color{blue}92}} \\
& \tiny{Bampis et al. \cite{bampis2016encoding}} & \tiny{100} & \tiny{81.54} \\
& \tiny{Tsintotas et al. \cite{tsintotas2018assigning}} & \tiny{100} & \tiny{\textbf{\color{red}93.18}} \\
& \tiny{\textbf{FILD}} & \tiny{\textbf{100}} & \tiny{\textbf{\color{green}91.23}} \\
\hline
\multirow{2}{*}{\tiny{KITTI 05 \cite{fritsch2013new}}} & \tiny{Gehrig et al. \cite{gehrig2016visual}} & \tiny{100} & \tiny{\textbf{\color{blue}94}} \\
& \tiny{Bampis et al. \cite{bampis2016encoding}} & \tiny{100} & \tiny{84.80} \\
& \tiny{Tsintotas et al. \cite{tsintotas2018assigning}} & \tiny{100} & \tiny{\textbf{\color{red}94.20}} \\
& \tiny{\textbf{FILD}} & \tiny{\textbf{100}} & \tiny{\textbf{\color{green}85.15}} \\
\hline
\multirow{2}{*}{\tiny{Malaga 2009 Parking 6L \cite{blanco2009collection}}} & \tiny{G$\rm \acute{a}$lvez-L$\rm \acute{o}$pez et al. \cite{galvez2012bags}} & \tiny{100} & \tiny{74.75} \\
& \tiny{FAB-MAP 2.0 \cite{cummins2011appearance}} & \tiny{100} & \tiny{68.52}  \\
& \tiny{Bampis et al. \cite{bampis2016encoding}} & \tiny{100} & \tiny{76.78}  \\
& \tiny{IBuILD \cite{khan2015ibuild}} & \tiny{100} & \tiny{\textbf{\color{green}78.13}} \\
& \tiny{Tsintotas et al. \cite{tsintotas2018assigning}} & \tiny{100} & \tiny{\textbf{\color{red}87.99}}  \\
& \tiny{\textbf{FILD}} & \tiny{\textbf{100}} & \tiny{\textbf{\color{blue}80.54}} \\
\hline
\multirow{2}{*}{\tiny{New College \cite{smith2009new}}} & \tiny{G$\rm \acute{a}$lvez-L$\rm \acute{o}$pez et al. \cite{galvez2012bags}} & \tiny{100} & \tiny{55.92} \\
& \tiny{Bampis et al. \cite{bampis2016encoding}} & \tiny{100} & \tiny{\textbf{\color{green}77.55}}  \\
& \tiny{Tsintotas et al. \cite{tsintotas2018assigning}} & \tiny{100} & \tiny{\textbf{\color{blue}87.97}}  \\
& \tiny{\textbf{FILD}} & \tiny{\textbf{100}} & \tiny{\textbf{\color{red}89.94}} \\
\bottomrule
\end{tabular}
\end{center}
\end{table}

\subsection{Discussion}
The performance of our system depends on several factors: the classification accuracy of the CNN model, the retrieval precision and recall of the HNSW graphs, and the
effectiveness of the geometrical verification. In this case, the CNN features were extracted using the final average pooling layer of MobileNetV2. An increase in the classification accuracy will lead to an increase of recall in the whole LCD system. For example, we tested our system using the ResNet152 model provided by the author in \cite{zhou2018places}. The recall at 100\% precision for the New College dataset was 93.85\%, which was higher than our result of 89.94\%. The reason why we have not used ResNet152 was that it cost more time in forwarding time, about 135 ms in GPU, which is intolerable for mobile robot applications. In the future, we will try to improve the classification accuracy using the Place365 \cite{zhou2018places} dataset. The performance of different parameters of the HNSW graphs were exhaustively evaluated. However, we did not fully utilize the similarity scores of the query and the returned images. A proper threshold may have helped us eliminate false positives. In the geometrical verification step, the hashing bits $m$ and the ratio $\varepsilon$ are important for the recall and the processing time. We plan to accelerate the CasHash \cite{cheng2014fast} algorithm using hardware instruction set or optimized math functions, which should enable us to use more bits to achieve higher recall with suitable time costs.


\section{CONCLUSIONS}

In this paper, an online, incremental approach for fast loop closure detection is presented. The proposed method is based on the GPU computed features and HNSW graph vocabulary construction. A novel geometrical verification method based on hashing codes is introduced, which is coupled with binary ratio test to generate loop closure. The approach is evaluated on different publicly available outdoor datasets, and the results show that it achieve fairly good results compared with other state-of-the-art methods, which is capable of generating higher recall at 100\% precision.

\section{ACKNOWLEDGMENTS}
The authors would like to thank Dr. Konstantinos A. Tsintotas for kindly offering GT information for the datasets, and Dr. Cong Leng for the constructive suggestion.

\bibliographystyle{ieeetran}
\bibliography{bible}

\end{document}